\documentclass[a4paper,fleqn]{cas-dc}

\usepackage{graphicx}
\graphicspath{{./figs/}}
\DeclareGraphicsExtensions{.pdf,.jpeg,.png,.jpg}

\usepackage[numbers]{natbib}
\usepackage{makecell}

\usepackage{balance}
\usepackage{amsmath}
\usepackage{amssymb}
\usepackage{flushend}
\usepackage{microtype}
\usepackage{footmisc}

\usepackage{algorithm}
\usepackage{algorithmic}
\usepackage{color}
\usepackage{adjustbox}
\usepackage{multirow}
\usepackage{booktabs}
\usepackage{amssymb}
\usepackage{listings} 
\usepackage{placeins} 
\usepackage{dialogue}
\usepackage{tcolorbox}

\newtcolorbox{speakerbox}[2][]{
    colback=gray!10,        
    colframe=black,         
    colbacktitle=gray!20,   
    coltitle=black,         
    fonttitle=\bfseries,    
    title={#2},             
    left=2mm,
    right=2mm,
    top=2mm,
    bottom=2mm,
    arc=2mm,
    boxrule=0.5pt,
    #1
}

\def\tsc#1{\csdef{#1}{\textsc{\lowercase{#1}}\xspace}}
\tsc{NER}
\tsc{LLM}
\tsc{SC}
\DeclareMathOperator*{\argmax}{arg\,max}

\begin{document}
\let\WriteBookmarks\relax
\def\floatpagepagefraction{1}
\def\textpagefraction{.001}
 \let\printorcid\relax
 
\shorttitle{ReverseNER: Enabling Large Language Models to Self-Hint for Named Entity Recognition via the Reversed Process}    

\shortauthors{Anbang Wang et al}  

\title [mode = title]{ReverseNER: Enabling Large Language Models to Self-Hint for Named Entity Recognition via the Reversed Process}  

\author[1]{Anbang Wang}
\author[2]{Difei Mei}
\author[3]{Zhichao Zhang}
\author[1]{Xiuxiu Bai}
\cortext[cor1]{Corresponding author}
\cormark[1]
\ead{xiubai@xjtu.edu.cn}

\author[1]{Ran Yao}
\author[1]{Zewen Fang}
\author[2]{Min Hu}
\author[2]{Zhirui Cao}
\author[2]{Haitao Sun}
\author[2]{Yifeng Guo}
\author[2]{Hongyao Zhou}
\author[2]{Yu Guo}

\address[1]{School of Software Engineering, Xi'an Jiaotong University, Xi'an, 710049, China }
\address[2]{China Mobile System Integration Co.,Ltd., Shijiazhuang, 050011, China}
\address[3]{China Mobile Communications Group Co.,Ltd., Beijing, 100033, China}

\begin{abstract}
This paper presents ReverseNER, a method aimed at overcoming the limitation of large language models (LLMs) in zero-shot named entity recognition (NER) tasks, arising from their reliance on pre-provided demonstrations. ReverseNER tackles this challenge by constructing a reliable example library composed of dozens of entity-labeled sentences, generated through the reverse process of NER. Specifically, while conventional NER methods label entities in a sentence, ReverseNER features reversing the process by using an LLM to generate entities from their definitions and subsequently expand them into full sentences. During the entity expansion process, the LLM is guided to generate sentences by replicating the structures of a set of specific \textsl{feature sentences}, extracted from the task sentences by clustering. This expansion process produces dozens of entity-labeled task-relevant sentences. After constructing the example library, the method selects several semantically similar entity-labeled examples for each task sentence as references to facilitate the LLM's entity recognition. We also propose an entity-level self-consistency scoring mechanism to improve NER performance with LLMs. Experiments show that ReverseNER significantly outperforms other zero-shot NER methods with LLMs, marking a notable improvement in NER for domains without labeled data, while declining computational resource consumption.
\end{abstract}





\maketitle

\section{Introduction}

The advent of large language models (LLMs) has marked a significant technological leap in natural language processing (NLP). Owing to extensive pre-training and parameter optimization, LLMs such as GPT-3~\cite{10.5555/3495724.3495883} have demonstrated exceptional performance across various NLP tasks. By capturing semantics, syntax, and contextual relationships from vast corpora, these models have not only achieved breakthroughs in traditional tasks like machine translation~\cite{Tang2020MultilingualTW}, relation extraction~\cite{Wadhwa2023-fk}, knowledge graphs manufacture~\cite{ANADIOTIS2022101846} and named entity recognition (NER)~\cite{Wang2023GPTNER}, but also exhibited strong generalization capabilities in generative tasks such as text generation~\cite{yu2023generateretrievelargelanguage}, code evaluation~\cite{MOHAMED2025102473} and dialogue generation~\cite{zheng-etal-2023-augesc}.

Despite the impressive performance of LLMs across various tasks, they still encounter challenges in zero-shot NER, which requires the models to identify entity types without prior exposure to annotated data. In contrast, traditional NER methods typically rely on extensively annotated datasets for supervised learning \cite{10.5555/645530.655813}. Although LLMs exhibit some transfer learning capabilities, they continue to struggle with complex NER tasks in zero-shot settings due to insufficient generalization and an imprecise understanding of domain-specific entities. This limitation stem from the model's dependence on the diversity and richness of its training corpus for recognizing entity contexts. When confronted with new entity types or contexts, LLMs often fail to match entity categories accurately. 

Previous works on zero-shot NER using LLMs have primarily focused on refining prompt templates to enhance performance~\cite{Hu2024-hx}\cite{xie2023empiricalstudyzeroshotner}\cite{zamai2024lessinstructmoreenriching}, but the results have yet to yield significant improvements. Another study explores decomposing the identification of different entity types into multiple independent Q\&A tasks to achieve more accurate results~\cite{xie2023empiricalstudyzeroshotner}. While this approach provides limited improvements, it also significantly increases computational resource consumption due to the increased number of Q\&A tasks.

\begin{figure}
	\centering
	\includegraphics[width=.9\columnwidth]{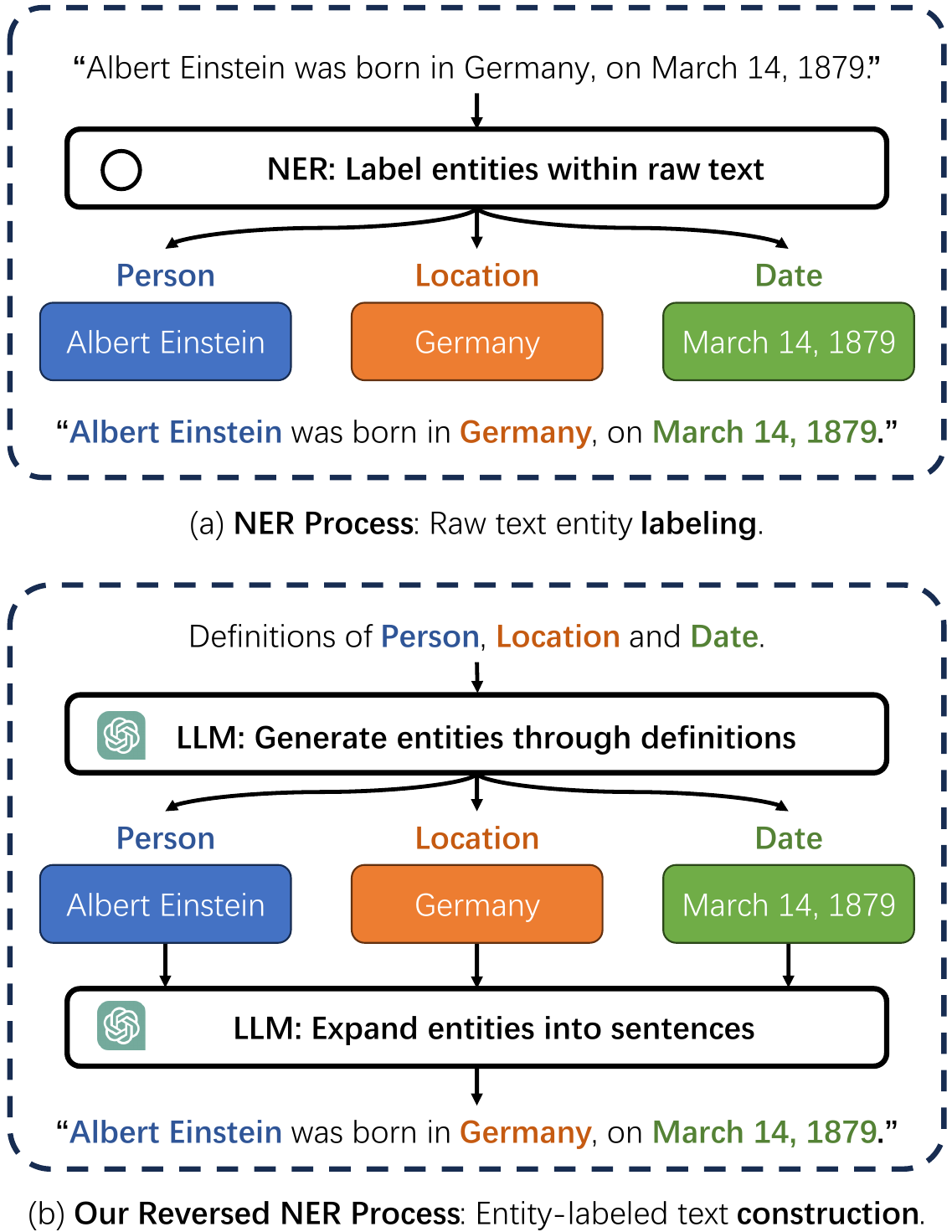}
	\caption{The processes of NER and Reversed NER. Both of them output entity-labeled sentences. (a) NER Process takes raw texts as input and outputs the result through entity labeling; (b) Our Reversed NER Process takes definitions of entities as the only input and outputs the result through generating entities and expanding entities into sentences. The Reversed NER Process requires a generative LLM to work.}
	\label{NER_and_ReverseNER}
\end{figure}

Given that LLMs perform much better in few-shot settings \cite{Ashok2023PromptNERPF}, we propose addressing the limitation of LLMs in zero-shot NER tasks by converting the zero-shot approach into a "pseudo few-shot" approach. Inspired by this, we introduce a novel method called ReverseNER, which enables the LLM to construct an example library by reversing the NER process (Figure~\ref{NER_and_ReverseNER}) to provide self-hints. In the reversed NER process, the method takes only the definitions of each entity as input and generates entities that match these definitions. The LLM is then instructed to expand these entities into full sentences. Sentences constructed through entity expansion are inherently labeled with given entities, as a result, these entity-labeled sentences are qualified as examples to guide the LLM during the actual NER procedure. 

We also present the method of feature sentence extraction through clustering to save computational resources while ensuring the task-relevance and diversity of the example library. Tested on four datasets, our method outperforms the baseline by 5.43 to 7.69 micro F1 and surpasses existing methods with lower computational resource consumption on two public English datasets, CoNLL03~\cite{tjong-kim-sang-de-meulder-2003-introduction} and WikiGold~\cite{balasuriya-etal-2009-named}.

The contributions of this paper are as follows:

\begin{enumerate} 
    \item A novel method for constructing accurate NER examples by reversing the conventional NER process. 
    \item Clustering techniques applied to the task dataset to decline computational resource consumption while ensuring task relevance and diversity in the constructed examples. 
    \item An entity-level self-consistency scoring mechanism designed to enhance precision and improve the micro F1, especially when increased computational cost can be justified by performance gains.
    \item Superior performance in micro F1 on public datasets compared to existing zero-shot methods using LLMs. 
\end{enumerate}

\section{Related work}

\subsection{Named Entity Recognition}

Named entity recognition is a fundamental task in NLP. Early approaches, such as Lee et al.~\cite{LEE2007575}, rely on bootstrapping and contextual pattern learning for fine-grained geographic entity annotation. They further introduce location normalization to resolve ambiguities like distinguishing cities from states, demonstrating the effectiveness of combining patterns with gazetteers. With the advent of machine learning, research has shifted toward sequence labeling formulations for NER tasks. For instance, Hammerton et al.~\cite{hammerton-2003-named} utilize unidirectional LSTMs~\cite{hochreiter1997long} to obtain token-level representations, which are then fed into a softmax classifier for entity classification; Collobert et al.~\cite{10.5555/1953048.2078186} employ CNNs to embed each input word and used Conditional Random Fields (CRFs) to decode each embedding into entity labels; Li et al.~\cite{li-etal-2020-unified} formulate the NER task as a Machine Reading Comprehension (MRC) task and further optimizes the performance of the MRC model using the Dice Loss function; Asghari et al.~\cite{ASGHARI2022184} propose BINER, a low-cost biomedical NER model that combines Bidirectional LSTM and CFRs, achieving competitive performance with limited resources.

After the advent of Transformer-based models~\cite{Attention} like BERT (Bidirectional Encoder Representations from Transformers)~\cite{devlin-etal-2019-bert}, many researchers have conducted NER studies using BERT. One of the earliest advancements is introduced by Devlin et al.~\cite{devlin-etal-2019-bert} with the BERT model, where fine-tuning on specific NER datasets significantly improves performance due to BERT's ability to capture bidirectional contextual information, which is crucial for understanding entity boundaries in text. More recently, Bartolini et al.~\cite{BARTOLINI2023102291} propose COSINER, a context similarity-based data augmentation method that generates plausible augmented samples for NER tasks. By replacing entity mentions with contextually similar ones, COSINER effectively improves model performance, particularly in low-resource and few-shot scenarios, while reducing the noise introduced by traditional augmentation techniques.

\subsection{Applications of LLMs in NER}

In recent years, explorations have been made of transforming NER into a generative task using LLMs. Wang et al.~\cite{Wang2023GPTNER} propose GPT-NER, a method that takes raw sentences as the inputs and outputs sentences with labeled entities by leveraging GPT models, rather than relying on the traditional sequence labeling approach. This shift to generation-based approaches enables more flexible in-context learning, making it particularly advantageous in scenarios with limited labeled data. Furthermore, GPT-NER employs a method of selecting relevant examples by calculating the similarity between the current input and a set of candidate examples. Although only a few examples are provided to the LLM during each recognition task, the method benefits from exposing the entire training set to the LLM to select relevant examples. As a result, in both true few-shot and zero-shot settings, GPT-NER demonstrates significantly weaker performance.

True few-shot and zero-shot NER models aim to generalize with little or no labeled data, making the task considerably more challenging for the models. Unlike GPT-NER, these approaches must rely on extra mechanisms, such as meta-learning or prompt engineering, to perform well in scenarios where labeled data is scarce or absent. Cheng et al.~\cite{CHENG2024100099} enhance few-shot NER by structuring prompts into three components: task definition, demonstration, and format. They also integrate feedback and error analysis to improve overall performance. Ashok et al.~\cite{Ashok2023PromptNERPF} introduce PromptNER, an approach that utilizes cloze-style prompts and entity definitions to leverage the pretraining capabilities of LLMs for recognizing entities with minimal task-specific annotations. This method claims to be the state-of-the-art method in the field of few-shot NER with pre-trained LLMs. However, its performance drops significantly in zero-shot scenarios.

Most recently, Xie et al.~\cite{xie2024selfimproving} introduce a self-improving framework for zero-shot NER. Their approach involves randomly selecting $500$ sentences from the dataset, which are then processed through vanilla zero-shot NER using an LLM. In this context, vanilla zero-shot NER refers to directly providing the sentences and the task description to the LLM, which then returns the recognized entities as its output. Each sentence undergoes recognition five times, and results with low self-consistency scores are filtered out using a majority voting mechanism, thereby creating a more reliable, LLM-generated pseudo-labeled sentence set. This refined set is used as an example library to hint at the LLM during inference for entity recognition. If needed, the resulting example library can be used as the input for the LLM to generate a more reliable next-generation example library. This mechanism enhances zero-shot performance on NER tasks across several benchmarks, including CoNLL03~\cite{tjong-kim-sang-de-meulder-2003-introduction}, WikiGold~\cite{balasuriya-etal-2009-named} and other datasets. However, this method requires heavy additional LLM invocations to generate enough entities ($500\times 5$ additional invocations without the self-improving iterations, $500\times 5 \times it$ additional invocations if performing $it$ iterations of the self-improving mechanism). And since the self-consistency mechanism can only mitigate errors caused by LLM hallucinations, it cannot detect or filter errors arising from the LLM's limitation in context understanding. As a result, the reliability of the generated examples remains questionable. In contrast, our method, which constructs an example library by reversing the conventional NER process, not only ensures the reliability of the example library but also consumes much less computational resources.

\section{Method}
Our insight lies in generating and providing LLMs with accurate and task-relevant examples to boost their performance in NER tasks. To achieve this in zero-shot scenarios with minimal LLM invocations, we propose the ReverseNER method, which features constructing an example library by reversing the conventional NER process. Specifically, our approach first generates entities based on definitions of each entity type, and subsequently prompts the LLM to expand these entities into complete sentences. During this sentence generation phase, a subset of sentences from the task set will be provided as structural and semantic references to guide the model. This example library aims to further enhance the model's performance in subsequent NER tasks by offering task-specific, high-quality examples.

\begin{figure*}
	\centering
	\includegraphics[width=.90\textwidth]{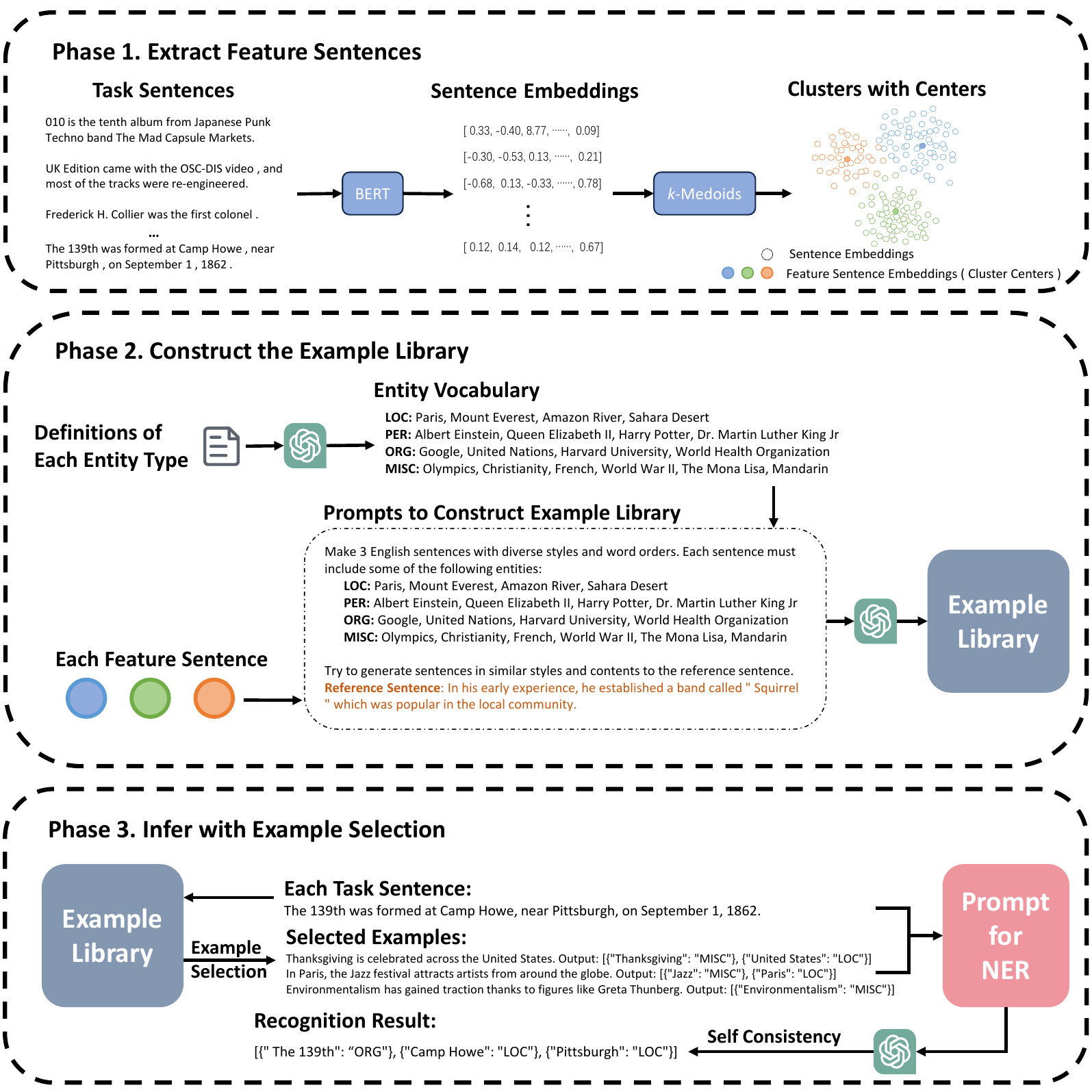}
	\caption{Overview of the entire ReverseNER Process: Clustering, Constructing Entity-labeled Exapmles and Selecting .}
	\label{FIG:1}
\end{figure*}

The main workflow, shown in Figure~\ref{FIG:1}, has three main phases: extracting \textsl{feature sentences} from task sets, constructing an example library with LLM, and for each task sentence, selecting the few most semantically similar entity-labeled examples and incorporating them into the prompt as examples.

\subsection{Feature Sentence Extraction} \label{FSE}
In this phase, the input consists of the entire task set, which includes many raw sentences, while the output is a set of \textsl{feature sentences} that are semantically similar to the majority of the sentences in the task set. This approach aims to ensure that, during the subsequent construction of the example library, the sentences in the example library remain semantically similar to those in the task set, while minimizing the additional usage of LLMs.

\subsubsection{Sentence Embedding Using BERT} The BERT model is employed to map sentences from the task set into high-dimensional vector representations. BERT is a pre-trained model based on the Transformer architecture, which effectively captures contextual information through a bidirectional encoder, generating deep, high-dimensional semantic embeddings for each sentence~\cite{devlin-etal-2019-bert}. Compared to traditional methods such as the Bag-of-Words model~\cite{bag-of-words} and TF-IDF~\cite{TF-IDF}, BERT captures the semantic similarity between texts more effectively. Specifically, given a sentence $S = [w_1, w_2, \ldots, w_\text{end}]$ composed of words $w_i$, BERT generates the embedding vector of the sentence $v_s$ denoted as:
\begin{equation}
v_s = f(H) = f(\text{BERT}(S)),
\end{equation}
where $H$ denotes the hidden state matrix generated by BERT, and $f$ denotes a post-processing function, which transforms these hidden states into a fixed-length vector that represents the entire sentence.

These generated embeddings enable the clustering of sentences, which in turn facilitates the identification of representative sentences that can be used to generate high-quality reference examples.

\subsubsection{Task Set Clustering} After obtaining sentence embeddings with BERT, the $k$-Medoids clustering algorithm is applied for sentence clustering. Unlike $k$-Means, $k$-Medoids selects \textsl{actual data points} as cluster centers. We select the sentences corresponding to these cluster centers as \textsl{feature sentences}. $k$-Medoids is also more robust to noise and outliers. Since it takes actual sentences as cluster centers, it is less affected by irrelevant sentences compared to $k$-Means, which is sensitive to outliers due to its use of average vectors~\cite{kaufman1987clustering}. 

Furthermore, $k$-Medoids can integrate \textsl{cosine similarity} instead of Euclidean distance. Cosine similarity measures the directional relationship between sentence embeddings, making it more suitable for capturing semantic similarity in text, as it focuses on the relative positioning of sentences rather than their norms~\cite{manning1999foundations}. The cosine similarity between two vectors $a$ and $b$ is defined as:

\begin{equation}
\text{sim}(a, b) = \cos(a, b) = \frac{a \cdot b}{\|a\|\|b\|}
\end{equation}
where $a \cdot b$ is the dot product of the vectors $a$ and $b$, and $\|a\|$ and $\|b\|$ are the norms of the vectors $a$ and $b$, respectively.

The specific steps of the $k$-Medoids clustering algorithm are as follows:

\begin{itemize}
    \item \textbf{Initialize Medoids}: Randomly select $k$ embeddings from the set of embedding vectors as the initial medoids.
    
    \item \textbf{Assign Embeddings to Medoids}: Measure the similarity between sentence embeddings using cosine similarity. For each embedding $v_i$, identify its nearest medoid $c_j$ and assign $v_i$ to the cluster $C_j$ corresponding to $c_j$. Specifically, for each $v_i$, we select the medoid $c_j$ that maximizes the cosine similarity:
    \begin{equation}
        c_j = \argmax_{c_j \in \mathcal{M}} \text{sim}(v_i, c_j),
    \end{equation}

    where $\mathcal{M}$ denotes the set of medoids.
    
    \item \textbf{Update Medoids}: For each cluster $C_j$, find the embedding that maximizes the total cosine similarity to all other embeddings in the cluster and designate it as the new medoid, formulated as:
    \begin{equation}
        c_j = \argmax_{v \in C_j} \sum_{v_i \in C_j} \text{sim}(v, v_i).
    \end{equation}

    In other words, extract a representative embedding within the cluster such that the total cosine similarity between this embedding and all other embeddings in the cluster is maximized.
    
    \item \textbf{Iterate}: Repeat the steps of assigning embeddings and updating medoids until the cluster centers no longer change or reach the maximum number of iterations.
\end{itemize}

The use of $k$-Medoids facilitates the extraction of \textsl{feature sentences} that are both representative and diverse. By selecting the cluster centers that capture different aspects of the dataset, we can ensure that the example library includes a wide range of sentence structures and contexts. Such diversity is crucial for the LLMs to handle a variety of linguistic patterns and entities, ultimately leading to improved generalization and performance.

\subsection{Example Library Construction}\label{ELC}

NER algorithms take task sentences as the input and output recognized entities. However, especially in zero-shot scenarios, these recognized results do not qualify as gold labels, which comes from the imperfect semantic understanding of the models, and consequently, example libraries constructed from these results may lack reliability. On the contrary, if the generation algorithms take entities as input and output a sentence that includes these entities, i.e., the reverse process of NER, the result will be much more reliable. Inspired by this, we present a novel method to construct the example library based on such reversed process.

Specifically, the example library construction in the ReverseNER method involves two stages using LLMs: first, generating a set of entities based on their definitions, then expanding these entities into sentences.

\subsubsection{Entity Vocabulary Generation}\label{EVG}
In this stage, the input consists of the entity type set and the corresponding definitions for all entities, while the output is the entity vocabulary—a list of entities mapped to their respective types.

The entity types and their definitions are first provided to an LLM, and then the LLM generates a set of specific entities based on the definitions for each type. The LLM is instructed to ensure diversity in the generated entities. The dialogue demonstration of the entity vocabulary generation process is shown below:

\begin{speakerbox}{\begin{flushright}
Prompt\end{flushright}
}
Here is an entity type set: [PER, LOC, ORG, MISC].\\

The definition of PER is ...;

The definition of LOC is ...;

The definition of ORG is ....\\

Please imagine a list of at least 3 diverse words for each entity type in the set.
\end{speakerbox}

\begin{speakerbox}{LLM}
\textbf{PER: }John Smith, Marie Curie, Albert Einstein

\textbf{LOC: }Paris, Amazon Rainforest, Mount Everest

\textbf{ORG: }United Nations, Red Cross, NASA
\end{speakerbox}

\subsubsection{Entity-to-Sentence Expansion}
In this stage, the input is the generated entity vocabulary in Section~\ref{EVG} and \textsl{feature sentences} extracted in Section~\ref{FSE}, and the output is an example library comprising a collection of entity-labeled sentences.

For each feature sentence, the LLM expands the generated entity vocabulary into sentences, mimicking the structure and format of the given feature sentence during the expansion process. These newly generated sentences should maintain structural consistency with the original sentences and ensure the previously generated entity nouns are included. Through this process, entity-labeled examples, are produced, ultimately constructing a fully annotated example library to support subsequent NER tasks. The dialogue demonstration of the entity vocabulary generation step is shown as follows:

\begin{speakerbox}{\begin{flushright}
Prompt\end{flushright}
}
Here is an entity type set: [PER, LOC, ORG].\\

Construct a sentence that contains entities from the following words:\\
PER: John Smith, Marie Curie, Albert Einstein\\
LOC: Paris, Amazon Rainforest, Mount Everest\\
ORG: United Nations, Red Cross, NASA\\

Output the sentence along with the utilized entities.\\
Here is a reference sentence, try to mimic its style and content: John and Sarah work for UNICEF in New York.
\end{speakerbox}

\begin{speakerbox}{LLM}
\textbf{Sentence:} John Smith and Marie Curie work for NASA in Paris.\\
\textbf{Entities:} [\{"John Smith":"PER"\},\\
\{"Marie Curie":"PER"\},\\
\{"NASA":"ORG"\}, \{"Paris":"LOC"\}]
\end{speakerbox}
As shown in the dialogue above, the entity-labeled examples generated through the reversed NER process may not always be semantically realistic. Still, the entities in the examples are correctly labeled, ensuring reliability as valid examples for NER tasks.

\subsection{Example-Driven Inference}
In this phase, the input consists of the full task set and the example library constructed in Section~\ref{ELC}, while the output is the final NER result, with each task sentence annotated with its corresponding entities.

\subsubsection{Example Selection} To identify the most relevant entity-labeled examples for each task sentence, a similarity matrix must be calculated between the sentences in the example library and those in the task set. To achieve this, we first use the BERT model to map each sentence in the example library into high-dimensional embedding vectors. Then, for each sentence in the task set, the cosine similarity between it and every sentence in the example library is calculated, resulting in a similarity matrix of size $N_{\mathcal{T}} \times N_{\mathcal{E}}$, where $N_{\mathcal{T}}$ denotes the number of sentences in the task set, and $N_{\mathcal{E}}$ denotes the number of sentences in the example library. 

After constructing the similarity matrix, for each task sentence, we select the entity-labeled examples with the highest cosine similarity. Specifically, for a task sentence $t_i$ and an example sentence $e_j$, we find $K$ example sentences with the highest cosine similarity:
\begin{equation}
    \hat{\mathbf{e}}_i = \underset{\mathbf{e}_j \in \mathcal{E}}{\text{arg top-$K$}} \, \text{sim}(t_i, e_j),
\end{equation}

where $t_i$ is the embedding vector of the $i$-th task sentence, $\mathbf{e}_j$ is the embedding vector of the $j$-th example sentence from the example library embedding set $\mathcal{E}$, and $\hat{\mathbf{e}}_i$ is the set of selected examples set for $t_i$. The $\text{arg top-$K$}$ operation selects $K$ example sentences with the highest cosine similarity for each task sentence.

Based on this, we select the most relevant entity-labeled examples to include in the prompt provided to the LLM during entity recognition.

\subsubsection{Inference with LLMs}

The constructed prompt to perform the NER task on a certain task sentence includes the following four elements: the entity type set to be recognized, definitions of each entity type, selected entity-labeled examples, and the task sentence itself. The LLM outputs the recognized entities within the task sentence. The dialogue demonstration of an inference attempt is shown as follows:
\begin{speakerbox}{\begin{flushright}
Prompt\end{flushright}
}
Perform named entity recognition task on this entity type set: [PER, LOC, ORG].\\

The definition of PER is ...;

The definition of LOC is ...;

The definition of ORG is ....\\

Example 1:\\
Input Sentence: John Smith and Marie Curie work for NASA in Paris.\\
Output: [\{"John Smith":"PER"\},\\
\{"Marie Curie":"PER"\},\\
\{"NASA":"ORG"\}, \{"Paris":"LOC"\}]\\

Example 2 to Example 5: ...\\

Task Input Sentence: Lily and Mark collaborate at the World Health Organization in Geneva.\\
Your output:
\end{speakerbox}

\begin{speakerbox}{LLM}
[\{"Lily":"PER"\}, \{"Mark":"PER"\}, \{"World Health Organization":"ORG"\}, \{"Geneva":"LOC"\}]
\end{speakerbox}

\subsection{Entity-Level Self-Consistency Scoring}

To mitigate the impact of LLM hallucinations, we perform multiple inference attempts on each sentence and introduce a self-consistency (SC) scoring mechanism. Unlike traditional Response-Level SC methods~\cite{wang2023selfconsistencyimproveschainthought}, our Entity-Level SC Scoring method focuses on counting the occurrence of individual entities across all attempts. This process is \textbf{optional}, improving accuracy at the cost of increased computational resources.

We first count the occurrences of each recognized entity across all inference attempts, denoting this total count as the self-consistency score $SC(\mathbf{E}_i)$ for entity $\mathbf{E}_i$. Next, for each inference attempt, we compute the average SC score for all entities predicted in that attempt. Formally, if an attempt predicts $n$ entities $\{\mathbf{E}_1, \mathbf{E}_2, \ldots, \mathbf{E}_n\}$, its self-consistency score $SC_{\text{attempt}}$ is formulated as:

\begin{equation}
SC_{\text{attempt}} = 
\begin{cases}
\displaystyle \frac{1}{n} \sum_{i=1}^{n} SC(\mathbf{E}_i), & \text{if } n > 0, \\[1em]
0, & \text{if } n = 0,
\end{cases}
\end{equation}

In cases where most attempts predict at least an entity, we then select the single attempt with the highest $SC_{\text{attempt}}$ and use it as the final output. Additionally, to handle sentences that contain no entities, we apply a majority-based rule: if more than half of the attempts predict no entity, we label the sentence as “no entity.”

Entity-level SC is particularly beneficial when dealing with uncertain or ambiguous entities. Because each entity’s SC score depends on how frequently it appears across attempts, responses containing less certain entities (i.e., those that do not appear in all or most attempts) naturally receive lower average scores. Consequently, this approach sacrifices some recall (proportion of true entities identified) in favor of higher precision (proportion of predicted entities that are correct). Such precision-oriented selection is especially useful for categories like MISC in CoNLL03 and Wikigold, where the definitions of entities can be less clear or overlap with other labels.

\section{Experiments}
In this section, we experiment on four benchmarks to evaluate our ReverseNER method in comparison to multiple baselines.

\subsection{Setup}\label{exp}
\paragraph{LLM:}
We utilize the \textbf{GPT-4o mini} model as the LLM backbone to generate examples and perform the NER task to evaluate our method. During the self-consistency scoring process, the temperature parameter is set to 0.8, and the inference attempt is repeated five times.

\paragraph{Example Library Construction:}
For English texts, we utilize the BERT-base-uncased model \cite{devlin-etal-2019-bert} to derive sentence embeddings, while for Chinese texts, we use the BERT-base-Chinese model \cite{devlin-etal-2019-bert}. Sentences from the task set are clustered into \textbf{10 categories}, and for each cluster's central sentence i.e., the \textsl{feature sentences}, the LLM generates \textbf{3 examples} by mimicking the sentence structure. Thus, these 30 entity-labeled examples constitute the entire example library for a certain dataset.

\paragraph{Datasets:}
We use four datasets to evaluate the performance of the NER task. For English texts, we utilize the CoNLL03 dataset~\cite{tjong-kim-sang-de-meulder-2003-introduction} and WikiGold dataset~\cite{balasuriya-etal-2009-named}. For Chinese texts, we use the People's Daily dataset~\cite{cui-etal-2016-consensus} and have constructed a novel dataset called GovAff, which consists of sentences pertaining to routine administrative affairs. The GovAff dataset encompasses four types of entities: person, government agency, corporation, and miscellaneous.

\paragraph{Baselines and Comparison:}
The performance of the vanilla zero-shot method, which operates without any provided examples (relying solely on the task description, entity types, and their corresponding definitions—serves as the baseline for comparison). In addition, we perform a vanilla few-shot NER experiment using 10 pre-labeled examples and compare their results with those of the proposed method in this study.

Furthermore, we evaluate our approach against other zero-shot and few-shot methods leveraging LLMs, including the Self-improving Framework by Xie et al.~\cite{xie2024selfimproving} for zero-shot recognition and PromptNER~\cite{ashok2023promptnerpromptingnamedentity} for both zero-shot and few-shot recognition. Given that these methods are tested on \textbf{GPT-3.5}, we also adopt \textbf{GPT-3.5} as our LLM backbone for comparison. The results of the Self-improving Framework and PromptNER are directly taken from the best performances reported in their respective papers. Resource consumption is calculated based on the total number of LLM invocations, with $N_{\mathcal{T}}$ denoting the total number of sentences to be recognized in each task set.

\subsection{Overall Performance}
\subsubsection{Comparison with Baseline Methods}
The results presented in Table~\ref{overall} demonstrate the superiority of the ReverseNER approach. On all datasets, ReverseNER without self-consistency, i.e., ReverseNER w/o SC, consistently outperforms the baseline methods, highlighting the method's effectiveness in leveraging LLMs under zero-shot settings.

When incorporating self-consistency, ReverseNER with SC further boosts performance across all datasets. Notably, on WikiGold and GovAff, micro F1 of the method reaches 78.45 and 83.53, respectively, representing a significant improvement over baseline methods. A slight gain introduced by SC is also observed on People’s Daily, a dataset without ambiguous entities, where ReverseNER SC achieves 75.62 in micro F1, compared to 75.46 without SC.
\renewcommand{\arraystretch}{1.5}

\begin{table*}[width=1.8\linewidth,cols=6,pos=h]
\caption{The micro F1 of ReverseNER method and its baseline comparisons with \textbf{GPT-4o mini}.}\label{overall}

\begin{tabular*}{.9\linewidth}{@{} Lccccc@{} }
\toprule
Method & CoNLL03 & WikiGold  & People's Daily & GovAff\\
\midrule
Vanilla Zero-shot                    & 71.69 & 70.97 & 67.93 & 78.10\\
Vanilla Few-shot                         & 72.71 & 75.86 & 71.23 & 83.40\\
\textbf{ReverseNER Zero-shot, w/o SC}    & 76.19 & 76.98 & 75.46 & 82.46 &\\
\textbf{ReverseNER Zero-shot, SC}       & \textbf{77.78} & \textbf{78.45} & \textbf{75.62} & \textbf{83.53}\\
\bottomrule
\end{tabular*}
\end{table*}

\subsubsection{Comparison with Existing Methods}
Table~\ref{benchmark} presents a comparison of the micro F1 achieved by our proposed ReverseNER method against existing zero-shot and few-shot approaches on the CoNLL03 and WikiGold datasets, along with an analysis of resource consumption. Compared to the Self-improving Framework with 1 Iteration, ReverseNER w/o SC demonstrates superior performance while significantly reducing resource consumption. Specifically, for low-resource scenarios, ReverseNER w/o SC attains micro F1 of 75.63 on CoNLL03 and 74.98 on WikiGold, utilizing only $10+10^*+N_{\mathcal{T}}$ LLM invocations. In contrast, the Self-improving Framework with 1 Iteration achieves 74.99 and 73.98 on CoNLL03 and WikiGold, respectively, but requires $500 \times 5 + N_{\mathcal{T}}$ LLM invocations.

Furthermore, for high-resource zero-shot methods, our ReverseNER with SC surpasses the Self-improving Framework with 8 Iterations on the WikiGold dataset, although it does not outperform it on the CoNLL03 dataset. Nevertheless, ReverseNER consistently exhibits lower resource consumption compared to the Self-improving Framework. Notably, the performance of ReverseNER with SC on the WikiGold dataset even exceeds the theoretical upper bound of the Self-improving Framework.

\begin{table*}[width=1.8\linewidth,cols=6,pos=h]
\caption{The micro F1 and resource consumption of our ReverseNER method with its comparisons. All methods are evaluated using \textbf{GPT-3.5}. Resource consumption is measured by the total number of LLM invocations, where $N_{\mathcal{T}}$ represents the number of sentences in the task set. For the WikiGold dataset, $N_{\mathcal{T}}=1697$; for CoNLL03, $N_{\mathcal{T}}=3453$. Numbers or symbols marked with an asterisk (*) indicate that the output tokens required for this LLM invocation might significantly surpass the scope of a straightforward NER task.}\label{benchmark}

\begin{tabular*}{.9\linewidth}{@{} Lcccc@{} }
\toprule
\textbf{Method}  & \textbf{CoNLL03} & \textbf{WikiGold} & \textbf{Resource Consumption} \\
\noalign{\vskip 5pt}
\midrule
\multicolumn{4}{c}{\textbf{Low-Resource Zero-shot Methods}}\\
\noalign{\vskip 5pt}

Vanilla Zero-shot & 68.98 & 70.77 & $N_{\mathcal{T}}$\\
PromptNER, Zero-shot~\cite{ashok2023promptnerpromptingnamedentity} & 68.10 & - & $N_{\mathcal{T}}^*$\\
\noalign{\vskip 5pt}
\makecell[l]{Self-improving Framework~\cite{xie2024selfimproving},\\ 1 Iteration} & 74.99 & 73.98 & $500\times5+N_{\mathcal{T}}$\\
\noalign{\vskip 5pt}
\textbf{ReverseNER, w/o SC (Ours)} & \textbf{75.63} & \textbf{74.98} & $10+10^*+N_{\mathcal{T}}$\\

\noalign{\vskip 5pt}
\midrule
\multicolumn{4}{c}{\textbf{High-Resource Zero-shot Methods}}\\
\noalign{\vskip 5pt}

\makecell[l]{Self-improving Framework~\cite{xie2024selfimproving},\\ 8 Iterations} & \textbf{77.49} & 73.98 & $500\times5\times8+N_{\mathcal{T}}$\\
\noalign{\vskip 5pt}
\textbf{ReverseNER, SC (Ours)} & 76.77 & \textbf{76.72} & $10+10^*+5N_{\mathcal{T}}$\\
\noalign{\vskip 5pt}

\makecell[l]{Self-improving Framework~\cite{xie2024selfimproving},\\Theoretical  Upper Bound} & \textcolor{gray!80}{81.65} & \textcolor{gray!80}{76.64} & $+\infty$\\

\noalign{\vskip 8pt}
\midrule
\multicolumn{4}{c}{\textbf{Few-shot Methods}}\\
\noalign{\vskip 5pt}

Vanilla Few-shot & 71.55 & 72.32 & $N_{\mathcal{T}}$\\
PromptNER, Few-shot~\cite{ashok2023promptnerpromptingnamedentity} & 78.62 & - & $N_{\mathcal{T}}^*$\\
\bottomrule
\end{tabular*}
\end{table*}

\subsection{Analysis}
\subsubsection{Example Library Quality Evaluation}

To enhance the interpretability of our ReverseNER method, we use the \textbf{micro F1} metric to evaluate the correctness of entity labels across the entire example library. We also define two additional metrics to assess the quality of our example library: \textbf{Average of Highest Cosine Similarities (AHS)} and \textbf{Entity Diversity Ratio (EDR)}, focusing on similarity and diversity, respectively. Below are the definitions and explanations for each metric:

\textbf{Micro F1}: The micro F1 assesses the overall performance across all entity labels within the example library. It ranges from $0$ to $100$, with higher values indicating better performance. Since an LLM entirely generates the example library, this metric is obtained through \textsl{manual inspection} to identify any missing entities or labeling errors.

\textbf{Average of Highest Cosine Similarities (AHS)}: AHS measures the overall alignment between the input task set and the example library by averaging the highest cosine similarity scores for each sentence in the input dataset. It ranges from $0$ to $1$, where higher values signify greater similarity. Specifically, given a task set \( \mathcal{T} \) containing \( N_{\mathcal{T}} \) sentences $\{t_1, t_2, ...t_{N_{\mathcal{T}}}\}$ and an example library \( \mathcal{E} \), AHS is calculated as follows:
\begin{equation}
    \overline{\text{sim}}(t_i) = \max_{e_j \in \mathcal{E}} \, \text{sim}(t_i, e_j), \nonumber \\
\end{equation}
\begin{equation}
    \text{AHS} = \frac{1}{N_{\mathcal{T}}} \sum_{i=1}^{N_{\mathcal{T}}} \overline{\text{sim}}(t_i).
\end{equation}
This metric quantifies the overall alignment between the input and example datasets by assessing how closely each input sentence matches its most similar counterpart in the example library.

\textbf{Entity Diversity Ratio (EDR)}: EDR compares the semantic diversity of entities in the example library against that of the input dataset. It is defined for each entity type $\tau$ and ranges from $0$ to $\infty$, with values closer to $1$ indicating similar levels of diversity. The Entity Diversity (ED) for a given entity type of the example library or task set is defined as:
\begin{equation}
    \text{$ED(\tau$)} = \frac{2}{N_\tau(N_\tau-1)} \sum_{i=1}^{N-1} \sum_{j=i+1}^{N} \left( 1 - \text{sim}(\mathbf{E}_i, \mathbf{E}_j) \right),
\end{equation}
where \( \mathbf{E}_i \) and \( \mathbf{E}_j \) are two unique entities of the same type $\tau$ within the same example library or task set, and \( N_\tau \) is the total number of unique entities for that type.

The Entity Diversity Ratio (EDR) for entity type $\tau$ is then computed as the division of ED of the example library $\text{ED}_{\mathcal{E}}$ and the ED of the task set $\text{ED}_{\mathcal{T}}$:
\begin{equation}
    \text{EDR($\tau$)} = \frac{\text{ED}_{\mathcal{E}}(\tau)}{\text{ED}_{\mathcal{T}}(\tau)}.
\end{equation}

An EDR greater than 1 indicates that the example library has greater semantic diversity for the entity type compared to the task set, an EDR less than 1 means the task set has greater semantic diversity, and an EDR equal to 1 signifies identical levels of semantic diversity between the example library and the task set for that entity type.

Note that EDR is only about diversity. An EDR value being close to $1$ does not necessarily imply that the entities in the example library cover the same scope as those in the task set. Therefore, the EDR metric should be considered in conjunction with the AHS index for a more comprehensive evaluation.

The metrics of the constructed example library for each dataset in Section~\ref{exp} with \textbf{GPT-4o mini} are presented in Table~\ref{lib}. A key achievement is the correctness of the constructed example libraries, which all achieved a perfect micro F1 of 100.00. Given that each example library contains only 30 entity-labeled sentences, while the dataset contains thousands of sentences, the AHS metric values, ranging from 0.6934 to 0.7737, remain promising. Although the EDR metric is close to 1 for many entities, it shows less promise for some entities, such as MISC in the CoNLL03-based example library and GOV in the GovAff-based example library. The definitions of these entities are often ambiguous and encompass a broad semantic range.

\begin{table*}[width=1.8\linewidth,cols=7,pos=h]
\caption{Quality evaluation of constructed example library by correctness (micro F1), similarity to the dataset (AHS), and entity diversity ratio (EDR). Base Dataset means the example library is built for the certain dataset.}\label{lib}
\begin{tabular*}{.9\linewidth}{@{\extracolsep{\fill}} lcccccc @{}}
\toprule
 \textbf{Base Dataset}& \textbf{Micro F1} & \textbf{AHS} & \textbf{EDR 1}  & \textbf{EDR 2} & \textbf{EDR 3} & \textbf{EDR 4}\\
\midrule
CoNLL03        & 100.00 & 0.7737 & \makecell[c]{PER: 1.5956} & \makecell[c]{ORG: 1.0517} & \makecell[c]{LOC: 0.8346} & \makecell[c]{MISC: 0.7011} \\
WikiGold       & 100.00 & 0.7664 & \makecell[c]{PER: 1.5425} & \makecell[c]{ORG: 1.0784} & \makecell[c]{LOC: 0.9556} & \makecell[c]{MISC: 0.9822} \\
People's Daily & 100.00 & 0.6934 & \makecell[c]{PER: 1.3497} & \makecell[c]{ORG: 1.0699} & \makecell[c]{LOC: 1.3835} & - \\
GovAff         & 100.00 & 0.7107 & \makecell[c]{PER: 1.2124} & \makecell[c]{CORP: 0.9389} &  \makecell[c]{GOV: 0.6900} & \makecell[c]{MISC: 0.7674} \\

\bottomrule
\end{tabular*}
\end{table*}

\subsubsection{Detailed Performance}
Taking the WikiGold dataset as a case study, we conduct a detailed analysis of the recognition accuracy for each label. Results are shown in Table \ref{detailed_wikigold}. The WikiGold dataset comprises four types of entities: PER, ORG, LOC, and MISC, denoting persons, organizations, locations, and miscellaneous entities, respectively. Notably, the MISC label is more ambiguous in definition, covering entities that do not fall into the aforementioned categories but still qualify as proper nouns. Consequently, the recognition accuracy for this label tends to be suboptimal.

\begin{table*}[width=1.8\linewidth,cols=7,pos=h]
\centering
\caption{Detailed NER results on WikiGold dataset with \textbf{GPT-4o mini}. Methods with the best Precision for each metric are indicated with \underline{underlining}, those with the best Recall are \textsl{italicized}, and those with the highest F1 are highlighted in \textbf{bold}.}\label{detailed_wikigold}
\begin{tabular*}{.9\linewidth}{@{\extracolsep{\fill}} llccccc}
\toprule
\multirow{2}{*}{Method} & \multirow{2}{*}{Metric} & \multicolumn{4}{c}{Entities} & \multirow{2}{*}{Overall} \\
\cmidrule(lr){3-6}
& & PER & ORG & LOC & MISC & \\
\midrule
\multirow{3}{*}{Vanilla Zero-shot} 
& Precision & 87.14 & 70.91 & 83.33 & 32.73 & 63.97 \\
& Recall & \textsl{85.98} & \textsl{72.84} & 85.63 & 71.62 & \textsl{79.69} \\
& F1 & 86.56 & 71.86 & 84.46 & 44.93 & 70.97 \\
\midrule
\multirow{3}{*}{\shortstack[l]{Zero-shot\\ReverseNER, \\ w/o SC}} 
& Precision & 90.93 & 77.79 & \underline{87.75} & 47.86 & 74.96 \\
& Recall & 84.41 & 72.05 & 85.42 & \textsl{72.50} & 79.11 \\
& F1 & \textbf{87.55} & 74.81 & \textbf{86.57} & 57.66 & 76.98 \\
\midrule
\multirow{3}{*}{\shortstack[l]{Zero-shot\\ReverseNER, \\ SC}} 
& Precision & \underline{91.26} & \underline{80.78} & 85.69 & \underline{55.53} & \underline{78.43} \\
& Recall & 84.11 & 70.68 & \textsl{86.03} & 70.15 & 78.45 \\
& F1 & 87.54 & \textbf{75.39} & 85.86 & \textbf{61.99} & \textbf{78.45} \\
\bottomrule
\end{tabular*}
\end{table*}

By examining the Precision, Recall, and F1 scores of each label across different methods, we observe that the ReverseNER with SC voting consistently achieves the highest performance in terms of Precision and F1 scores for most labels, followed by the ReverseNER without SC voting. The vanilla zero-shot method, however, frequently achieves higher Recall, particularly for PER and ORG types. This phenomenon can be attributed to two primary factors: (1) The conservative strategy employed by the LLM when generating examples, where uncertain terms that may not belong to the current entity label are often excluded, leads to higher Precision but potentially lower Recall; (2) The SC voting algorithm tends to penalize outputs containing entities that are not consistently present across all outputs, even if they are correct in some cases, which further boosts Precision while reducing Recall. Consequently, these strategies improve the F1 score, particularly for labels with less clearly defined boundaries such as MISC categories, as they achieve a better balance between Precision and Recall.

\subsubsection{Impact of hyperparameters}
Our ReverseNER method involves two hyperparameters: the number of clusters used during feature sentence extraction and the number of entity-labeled examples that the LLM needs to construct per cluster. The product of these two hyperparameters determines the total number of labels in the example library. To evaluate the algorithm's stability under different parameter settings, we do not enable the self-consistency scoring mechanism and run each hyperparameter combination \textsl{20 times} with \textbf{GPT-4o mini}, calculating the mean and standard deviation.

Taking the WikiGold dataset as an example, Table~\ref{hyperparameters} illustrates the performance for each set of hyperparameter combinations. From the table, we can observe that keeping the total number of labels in the example library constant while increasing the number of clusters and decreasing the number of labels constructed per cluster leads to decreased standard deviation, standing for increased stability. However, for the F1 score, there is a peak point; either increasing or decreasing the number of clusters beyond this point negatively impacts the results.

The reason for this phenomenon may be that when the number of clusters is small, more labels need to be constructed for each cluster, resulting in a large number of labels with similar sentence structures in the example library. This makes it difficult for the LLM to receive diverse and contextually appropriate examples during each inferring, thereby reducing both the overall F1 score and stability. Conversely, when the number of clusters is too large, the sentence structures in the example library become more varied, but it also becomes more challenging to select the most contextually relevant examples for the current task sentence, leading to a slight decrease in the F1 score. 

\begin{figure}
	\centering
	\includegraphics[width=\columnwidth]{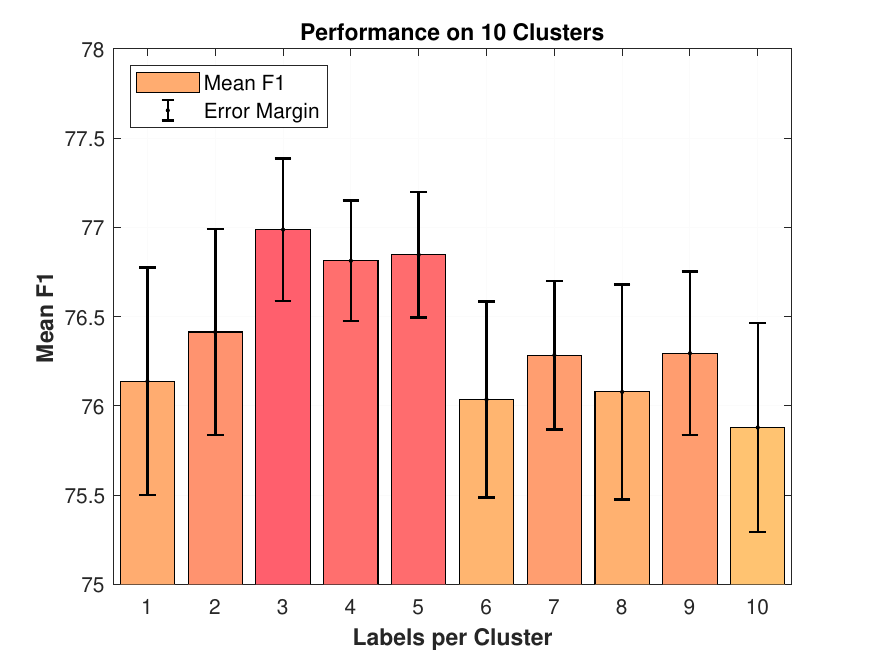}
	\caption{Average micro F1 on WikiGold when the number of clusters is fixed at 10, evaluated on \textbf{GPT-4o mini}.}
	\label{FIG:lalebs_per_cluster}
\end{figure}

Notably, as it is shown in Figure \ref{FIG:lalebs_per_cluster}, when the number of clusters is fixed, simply increasing the number of labels constructed for each cluster is unlikely to bring performance improvement and may instead lead to a decrease in F1 score stability.

\begin{table}[width=.9\linewidth,cols=3,pos=h!]
\centering
\caption{The micro F1 on WikiGold with different sets of hyperparameters, without self-consistency scoring, evaluated on \textbf{GPT-4o mini}.}\label{hyperparameters}
\begin{tabular}{cccl}
\toprule
Clusters & Labels per Cluster & Micro F1 \\
\midrule
1  & 30 & 75.60 $\pm$ 2.15 \\
2  & 15 & 75.72 $\pm$ 1.78 \\
3  & 10 & 76.02 $\pm$ 1.18 \\
5  & 6  & 76.07 $\pm$ 1.08 \\
6  & 5  & 76.71 $\pm$ 0.78 \\
10 & 3  & \textbf{76.98} $\pm$ 0.71 \\
15 & 2  & 76.28 $\pm$ 0.59 \\
30 & 1  & 76.19 $\pm$ 0.57 \\
\midrule
1  & 60 & 74.23 $\pm$ 3.60 \\
2  & 30 & 75.04 $\pm$ 1.43 \\
3  & 20 & 75.31 $\pm$ 1.29 \\
4  & 15 & 75.43 $\pm$ 1.21 \\
5  & 12 & 75.85 $\pm$ 1.18 \\
6  & 10 & 75.52 $\pm$ 0.96 \\
10 & 6  & 76.03 $\pm$ 0.89 \\
12 & 5  & 76.14 $\pm$ 0.95 \\
15 & 4  & 76.63 $\pm$ 0.86 \\
20 & 3  & 76.93 $\pm$ 0.78 \\
30 & 2  & 75.90 $\pm$ 0.73 \\
60 & 1  & 76.28 $\pm$ 0.54 \\
\bottomrule
\end{tabular}
\end{table}

\subsubsection{Effect of Entity-Level Self-Consistency Scoring}
The comparison presented in Table~\ref{scbenchmark} elucidates the performance differences between the Entity-Level Self-Consistency (SC) and the traditional Response-Level SC methods, which operate by selecting the most frequently occurring result from multiple iterations, across various datasets, evaluated using the F1 score metric. On the CoNLL03, WikiGold, and GovAff datasets, the Entity-Level SC method performs slightly better than the Response-Level SC method. In contrast, the Response-Level SC method outperforms the Entity-Level SC on the People’s Daily dataset.

This isolated underperformance for the Entity-Level SC can be attributed to the fact that there are only three clearly defined types of entities in the People's Daily Dataset: PER, LOC, and ORG, without the MISC type in other datasets, which does not have a well-defined boundary. The Entity-Level SC strategy textslasizes trading Recall for Precision to enhance the F1 score, but this approach is less effective in datasets where entity definitions are more straightforward, resulting in lower performance compared to the traditional Response-Level SC method.

\begin{table}[width=.9\linewidth,cols=3,pos=h]
\caption{The micro F1 comparison on different SC methods, evaluated on \textbf{GPT-4o mini}.}\label{scbenchmark}

\begin{tabular}{lccc}
\toprule
Dataset & Entity-Level SC & Response-Level SC \\
\midrule
CoNLL03 & \textbf{77.78} & 77.41\\
WikiGold & \textbf{78.45} & 78.00\\
People's Daily & 75.62 & \textbf{77.72}\\
GovAff & \textbf{83.53} & 83.34\\
\bottomrule
\end{tabular}
\end{table}
\subsubsection{Performance on Lightweight LLMs}
The performance of our method, evaluated on Qwen2.5-7b (qwen2.5-7b-instruct)~\cite{qwen25}, is shown in Table~\ref{qwen}. Our method provides a noticeable improvement in the zero-shot NER performance of Qwen2.5-7b, especially for the Chinese datasets, People's Daily and GovAff. This improvement can be attributed to the better performance of the Qwen series models on Chinese text and the higher quality of the constructed example library. Notably, the performance of Qwen2.5-7b on the People's Daily dataset is somewhat disappointing. One reason for this is that many sentences in the dataset triggered its safety mechanism, causing the LLM to refuse to provide any responses.

\begin{table}[width=.9\linewidth,cols=3,pos=h]
\caption{The micro F1 comparison on \textbf{Qwen2.5-7b}.}\label{qwen}

\begin{tabular}{lccc}
\toprule
Dataset & Vanilla Zero-shot & ReverseNER, w/o SC \\
\midrule
CoNLL03 & 60.36 & \textbf{62.46}\\
WikiGold & 63.88 & \textbf{66.70}\\
People's Daily & 52.37 & \textbf{57.02}\\
GovAff & 72.50 & \textbf{77.05}\\
\bottomrule
\end{tabular}
\end{table}

\subsubsection{Handling Isolated Zero-shot NER}
We define Isolated Zero-shot NER as the scenario in which each sentence is treated as an independent unit. In this setting, the model processes one sentence at a time, ensuring that only the information within that sentence is available for entity recognition. This contrasts with previous approaches where multiple sentences or entire documents might be processed together, potentially allowing context from surrounding sentences to influence the NER results.

Our ReverseNER can handle this challenge \textbf{without any modifications}. By treating each task sentence as a complete task set, the feature sentence extraction process in Section~\ref{FSE} reduces to selecting the sentence itself as the feature sentence. We then construct the example library by mimicking its structure and obtain 3 similar entity-labeled sentences as examples. Finally, we use these examples to guide the LLM during inference.

Taking WikiGold as a case study, we emulate an Isolated Zero-shot NER scenario on it. The results and computational resource consumption are shown in Table~\ref{iNER}. In the Isolated Zero-shot NER setting, our ReverseNER algorithm still demonstrates significant improvements over the baseline. In contrast, all other zero-shot methods used for comparison in previous sections simply cannot perform. However, when performing Isolated Zero-shot NER with our method, the need to construct an example library for each sentence increases the computational resource consumption.

\begin{table}[width=.9\linewidth,cols=3,pos=h]
\caption{The micro F1 and resource consumption of Isolated Zero-shot NER on WikiGold with \textbf{GPT-4o mini}. Resource consumption is measured by the total number of LLM invocations, $N_{\mathcal{T}}=1697$. Symbols marked with an asterisk (*) indicate that the output tokens required for this LLM invocation might significantly surpass the scope of a straightforward NER task.}\label{iNER}

\begin{tabular}{lccc}
\toprule
Method & micro F1 & Resource Consumption \\
\midrule
Vanilla Zero-shot & 70.97 & $N_{\mathcal{T}}$\\
ReverseNER, w/o SC & 76.22 & $2N_{\mathcal{T}}+N_{\mathcal{T}}^*$\\
ReverseNER, SC & 77.77 & $6N_{\mathcal{T}}+N_{\mathcal{T}}^*$\\
\bottomrule
\end{tabular}
\end{table}

\section{Conclusion}
This paper presents ReverseNER, a novel method designed to enhance zero-shot named entity recognition (NER) tasks by utilizing the generative capabilities of large language models (LLMs) in a self-generated, example-driven approach. 

Our experiments across multiple benchmarks demonstrate that ReverseNER consistently improves NER performance compared to other zero-shot methods with significantly lower resource consumption. Additionally, the incorporation of an entity-level self-consistency scoring mechanism further refines the precision of entity recognition, particularly in scenarios involving ambiguous entity types. The analysis highlights the effectiveness of our feature sentence extraction and example library construction processes, ensuring high-quality and diverse examples that significantly aid the LLM's performance. This paper also presents three metrics to evaluate the quality of the generated example library to increase interpretability.

Overall, ReverseNER contributes to the ongoing efforts to leverage LLMs for more effective and autonomous information extraction, providing a practical approach for improving zero-shot NER performance in various applications.

\section{Limitations and Future Work}
We acknowledge the following limitations of this study:
\begin{itemize}
    \item The effectiveness of the proposed method is highly dependent on the quality of manually provided entity type definitions.
    \item The proposed method may lead to a slight decline in Recall due to trade-offs between Precision and Recall.
    \item The optional self-consistency scoring mechanism increases the frequency of LLM invocations, which undermines the low-resource consumption advantage achieved through the reversed NER example library construction process.
\end{itemize}

This work primarily focuses on enhancing the performance of LLMs for NER tasks. However, the concept of guiding LLMs to self-hint through reversed processes holds the potential for improving performance in other NLP tasks, such as relation extraction, sentiment analysis, and semantic role labeling.

\section{Acknowledgment}
This work is funded by the Xi'an Jiaotong University-China Mobile Communications Group Co., Ltd. Digital Government Joint Institute.

\appendix
\section{Dataset Statistics}
Dataset statistics are shown in Table~\ref{Statistics}. For WikiGold and GovAff, we perform the zero-shot NER task on the entire dataset. For CoNLL03 and People's Daily, we restrict the zero-shot NER task to the official test sets. Since this work focuses on zero-shot NER, no training or validation data are utilized.
\begin{table}[width=.9\linewidth,cols=3,pos=h]
\caption{Dataset Statistics}\label{Statistics}

\begin{tabular}{lcc}
\toprule
Dataset & \makecell[c]{Sentences Tested\\in this Paper} & Full Dataset Size \\
\midrule
CoNLL03 & 3453 & 14382\\
WikiGold & 1697 & 1697\\
People's Daily & 3000 & 19484\\
GovAff & 4825 & 4825\\
\bottomrule
\end{tabular}
\end{table}

\section{Actual Prompts for Constructing the Example Library}

Taking the WikiGold dataset as an example, the prompt used for generating entity vocabularies of this dataset is illustrated in Listing~\ref{EntitiesGeneration}1. We repeat this prompt when constructing entity-labeled examples for each cluster.

The prompt for constructing entity-labeled examples in the example library is shown in Listing~\ref{LabelsGeneration}2. The entity vocabularies it uses are derived from the results of the corresponding entity vocabulary generation prompt. The reference sentence provided at the end corresponds to the sentence located at the center of the respective cluster. We repeat this prompt when constructing entity-labeled examples for each cluster. 

\begin{figure*}
\normalsize
\begin{minipage}{\textwidth}\label{EntitiesGeneration}
\begin{lstlisting}[caption={Entity Vocabulary Generation Prompt}]
Here is an entity type set: [PER, ORG, LOC, MISC].

Here are the explanations of each entity from the label set:
    PER means Person. Definition: Denotes individual people or fictional characters. This includes full names, nicknames, and titles when they are part of the name.
    ORG means Organization. Definition: Represents groups of people that are identified by a particular name. This includes companies, institutions, government bodies, agencies, and other formal organizations.
    LOC means Location. Definition: Refers to geographical entities such as countries, cities, landmarks, mountains, rivers, and any other physical locations.
    MISC means Miscellaneous. Definition: Covers entities that do not fall into the above categories but are still proper nouns. This includes nationalities, religions, events, languages, works of art, and other entities.

Please imagine a list of at least 6 diverse words for each entity type in the set. The output must be a JSON object, where keys are the entity types and values are lists of words. Here is the output JSON structure example you must follow:
{
    "PER": ["PER_1", "PER_2", "PER_3"],
    "ORG": ["ORG_1", "ORG_2", "ORG_3"],
    "LOC": ["LOC_1", "LOC_2", "LOC_3"],
    "MISC": ["MISC_1", "MISC_2", "MISC_3"]
}
\end{lstlisting}
\end{minipage}
\end{figure*}

\begin{figure*}[H] 
\centering
\normalsize
\begin{minipage}{\textwidth}\label{LabelsGeneration}
\begin{lstlisting}[caption={Entity-labeled Sentences Construction Prompt}]
Here is an entity label set: [PER, ORG, LOC, MISC], do as follows:
    Make 3 English sentences with diverse styles and word orders.
    Each sentence must contain one, two, or three entities from the following words:
        PER: Leonardo da Vinci, Nelson Mandela, Marie Curie, Frida Kahlo, Sherlock Holmes, Malala Yousafzai
        ORG: NASA, Amnesty International, Harvard University, World Wildlife Fund, Microsoft, International Monetary Fund
        LOC: Tokyo, Grand Canyon, Great Wall of China, Nile River, Mount Kilimanjaro, Sydney
        MISC: Jazz, Buddhism, Renaissance, Spanish, The Great Gatsby, Nobel Prize
    
If your generated sentences include entities that may belong to a certain type of entity label set but are not shown in the word list, mark them as well in the output JSON.
Output format:
[
    {
        "text": "Your sentence here.",
        "entities": [
            {"entity_text": "the PER entity","entity_label": "PER"},
            {"entity_text": "the ORG entity","entity_label": "ORG"},
            {"entity_text": "the LOC entity","entity_label": "LOC"},
            {"entity_text": "the MISC entity","entity_label": "MISC"}
        ]
    }
]

Here is a sentence for reference. You can reference its style and content, but ensure the generated sentence covers different topics and scenarios
Reference Sentence: He died in Hollywood, California.

\end{lstlisting}
\end{minipage}
\end{figure*}

\section{Actual Prompts for Performing NER on Each Task Input}

The prompt used for inferring a specific input sentence from the WikiGold dataset is shown in Listing~\ref{TestingPrompt}3. The provided examples are selected from the constructed example library based on cosine similarity.

\begin{figure*}[H]
\centering
\normalsize
\begin{minipage}{\textwidth}\label{TestingPrompt}
\begin{lstlisting}[caption={Final Recognition Prompt}]
Perform English NER task in the following entity type set: [PER, ORG, LOC, MISC]
Here are the explanations of each entity from the label set:
    PER means Person. Definition: Denotes individual people or fictional characters. This includes full names, nicknames, and titles when they are part of the name.
    ORG means Organization. Definition: Represents groups of people that are identified by a particular name. This includes companies, institutions, government bodies, agencies, and other formal organizations.
    LOC means Location. Definition: Refers to geographical entities such as countries, cities, landmarks, mountains, rivers, and any other physical locations.
    MISC means Miscellaneous. Definition: Covers entities that do not fall into the above categories but are still proper nouns. This includes nationalities, religions, events, languages, works of art, and other entities.
Output format: [{"entity 1 text": "entity 1 type"},{"entity 2 text": "entity 2 type"}]

Example 1: Input sentence: Marie Curie was awarded the Nobel Prize for her groundbreaking research in radioactivity. Output: [{"Marie Curie": "PER"}, {"Nobel Prize": "MISC"}]
Example 2: Input sentence: The World Cup is celebrated globally, bringing nations like Brazil and Germany together. Output: [{"World Cup": "MISC"}, {"Brazil": "LOC"}, {"Germany": "LOC"}]
Example 3: Input sentence: In 2019, Greenpeace launched a campaign focused on the Great Wall of China to raise awareness about environmental issues. Output: [{"Greenpeace": "ORG"}, {"Great Wall of China": "LOC"}]
Example 4: Input sentence: Nelson Mandela's advocacy for peace was often celebrated during Diwali festivities around the world. Output: [{"Nelson Mandela": "PER"}, {"Diwali": "MISC"}]
Example 5: Input sentence: Marie Curie's groundbreaking research in radiation opened new doors for the field of science at Harvard University. Output: [{"Marie Curie": "PER"}, {"Harvard University": "ORG"}]

Test Input Sentence: Another Nobel laureate scientist there was Richard J. Roberts. Output:

\end{lstlisting}
\end{minipage}
\end{figure*}

\FloatBarrier

\bibliographystyle{cas-model2-names}

\bibliography{cas-refs}




\end{document}